\documentclass[letterpaper, 10 pt, conference]{ieeeconf}  

\IEEEoverridecommandlockouts                              

\usepackage{graphics} 
\usepackage{epsfig} 
\usepackage{mathptmx} 
\usepackage{times} 
\usepackage{amsmath} 
\usepackage{amssymb}  
\usepackage{epsfig,subfigure}
\usepackage{bm} 
\newcommand{\vect}[1]{\boldsymbol{#1}}
\usepackage{soul} 
\usepackage{xcolor} 
 \usepackage[linesnumbered,ruled,vlined]{algorithm2e}
\sethlcolor{yellow}

\usepackage{rotating}
\usepackage{cite}
\usepackage{wrapfig}
\usepackage{float}
\usepackage{titlesec}
\usepackage{array}
\usepackage{booktabs}
\usepackage{gensymb}

\usepackage{makecell, cellspace}

\usepackage{hyperref}
\hypersetup{
  colorlinks=true,
  linkcolor=blue,
  filecolor=magenta,      
  urlcolor=cyan,
  citecolor=green,
}

 \SetKwInput{KwParam}{Param}

\title{\huge  
    DualShield: Safe Model Predictive Diffusion via Reachability Analysis for Interactive Autonomous Driving
} 
\author{ 
Rui Yang, Lei Zheng, Ruoyu Yao, and Jun Ma 
\thanks{
    Rui Yang, Lei Zheng, Ruoyu Yao, and Jun Ma are with the Robotics and Autonomous Systems Thrust, The Hong Kong University of Science and Technology (Guangzhou), Guangzhou 511453, China (e-mail: \{ryang253, lzheng135, ryang092\}@connect.hkust-gz.edu.cn; jun.ma@ust.hk).}
}  

\begin{document}

\maketitle
\thispagestyle{empty}
\pagestyle{empty}

\begin{abstract}

Diffusion models have emerged as a powerful approach for multimodal motion planning in autonomous driving. However, their practical deployment is typically hindered by the inherent difficulty in enforcing vehicle dynamics and a critical reliance on accurate predictions of other agents, making them prone to safety issues under uncertain interactions.
To address these limitations, we introduce \texttt{DualShield}, a planning and control framework that leverages Hamilton-Jacobi (HJ) reachability value functions in a dual capacity. First, the value functions act as proactive guidance, steering the diffusion denoising process towards safe and dynamically feasible regions. Second, they form a reactive safety shield using control barrier-value functions (CBVFs) to modify the executed actions and ensure safety.
This dual mechanism preserves the rich exploration capabilities of diffusion models while providing principled safety assurance under uncertain and even adversarial interactions. Simulations in challenging unprotected U-turn scenarios demonstrate that \texttt{DualShield} significantly improves both safety and task efficiency compared to leading methods from different planning paradigms under uncertainty.

\end{abstract}

\section{INTRODUCTION}
 The central challenge in autonomous driving is not merely to navigate efficiently, but to proactively handle complex interactions with unpredictable human drivers \cite{chen2022milestones, zheng2024}. This reality introduces a fundamental tension: on one hand, vehicles require the exploratory power of multimodal motion planning to act decisively, preventing behavioral freezing in complex scenarios \cite{zheng2024barrier}. On the other hand, they must adhere to strict, formal safety guarantees, even when faced with worst-case adversarial behaviors \cite{zhang2024interaction, carrizosa2024safe}. Therefore, developing a framework that unifies the flexibility of multimodal planning techniques with the non-negotiable assurance of formal safety is paramount for building truly trustworthy autonomous systems.


 In pursuit of this goal, two primary research directions have emerged. The first centers on optimization-based planners, exemplified by model predictive control (MPC). While valued for its receding horizon scheme and explicit handling of system dynamics and constraints \cite{camacho2007constrained, zheng2024spatiotemporal}, traditional gradient-based MPC struggles in real-world driving scenarios characterized by non-differentiable objective functions and non-convex constraints, often converging to suboptimal local minima \cite{zheng2022safe}. 
 To address this issue, sampling-based approaches like model predictive path integral (MPPI) have gained traction \cite{williams2018information, liu2025sampling}. By optimizing through stochastic sampling, MPPI can better handle non-convex problems.
 Nevertheless, it suffers from fundamental limitations in both safety and exploration. Its conventional reliance on soft penalties provides only fragile safety assurance. Even when augmented with formal safety layers \cite{yin2023shield, yin2022trajectory}, its core exploration mechanism based on local perturbations remains a performance bottleneck, resulting in high sensitivity to initialization \cite{trevisan2024biased, mohamed2025towards}. 

\begin{figure}[tb]
\begin{center}
\includegraphics[width=.95\columnwidth]{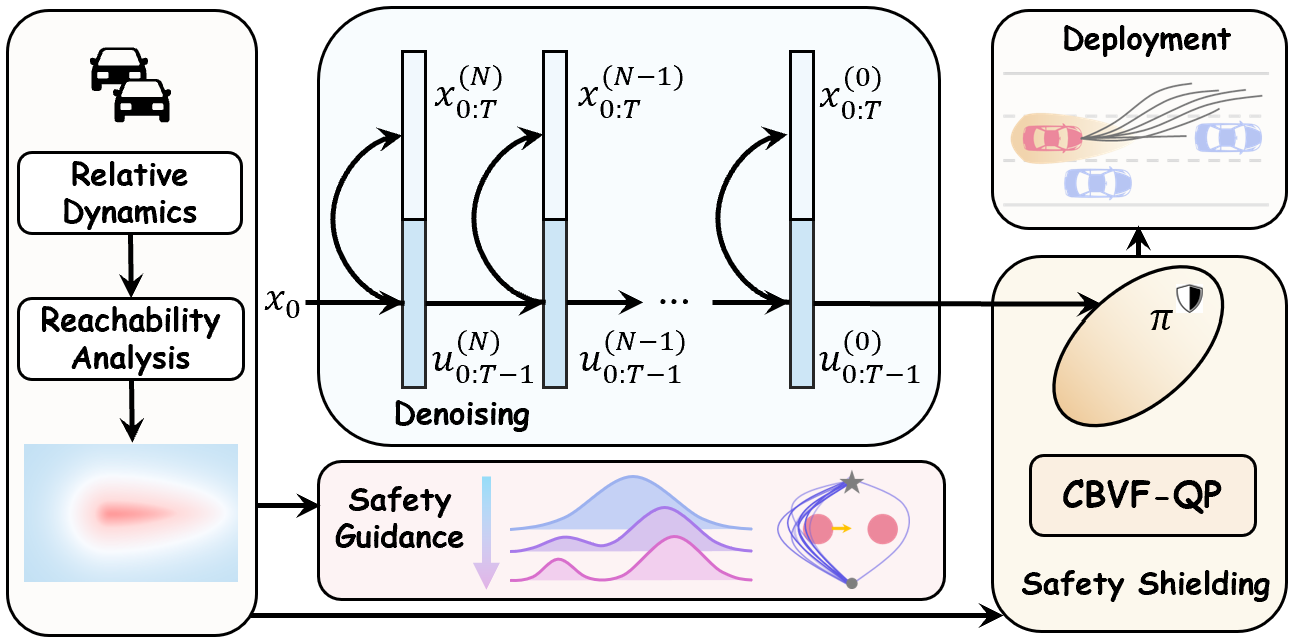}    \vspace{-1mm}
\caption{
Overview of the \texttt{DualShield} framework. The model-based diffusion generates candidate trajectories in a receding horizon scheme, leveraging a pre-computed HJ value function in a dual capacity: it proactively guides the denoising process away from high-risk regions, and serves as the final shield by forming a reactive CBVF-QP to filter the executed controls.
} \vspace{-6mm}
\label{fig:occ}
\end{center}
\end{figure}   


To address these exploration limitations, a second paradigm leverages generative models, particularly diffusion models  \cite{janner2022planning, zhong2023guided, mishra2023generative, jiang2023motiondiffuser}, for global exploration. Unlike local perturbations of MPPI, diffusion models learn a multimodal trajectory distribution,  enabling them to excel in highly non-convex problems by jointly optimizing discrete decisions and continuous trajectories  \cite{yang2024diffusion, yao2024calmm}. 
However, most existing approaches are trained purely from data,  failing to explicitly enforce hard system dynamics or safety constraints \cite{kondo2024cgd}.  



This limitation motivates recent model-based diffusion approaches, which reintegrate structured model priors into the denoising process, as the desired trajectory distribution is explicitly characterized by the objective function, constraints, and system dynamics \cite{pan2024model, jung2025joint}. These methods reframe the denoising process as a dynamics-aware trajectory generation by embedding the model components into the score function \cite{kurtz2025equality}. To address interaction-rich scenarios, a model predictive diffusion framework has been tailored for highway merging using Bayesian intent inference~\cite{knaup2025dual}. However, these methods still treat safety primarily through distance-based soft penalties, a strategy that lacks formal guarantees and is sensitive to hyperparameter tuning~\cite{yang2024diffusion, chen2024human, ma2025constraint}.

To achieve provable safety, control barrier function (CBF) defines a desired safe set as its zero-superlevel set and provides safety assurance by enforcing its forward invariance \cite{agrawal2017discrete, ames2019control, ames2014control}. Recent works have incorporated CBFs into diffusion frameworks by filtering generated trajectories through a safety layer in a quadratic programming (QP) formulation \cite{10995231} or embedding CBF constraints into the denoising score function \cite{10802549}. Nevertheless, designing effective CBFs is non-trivial, and they can be brittle under the profound uncertainty of interactive human vehicles (HVs) \cite{zhang2025automated, althoff2016set, koschi2020set}.
To explicitly account for multi-agent interactions, Hamilton–Jacobi (HJ) reachability analysis formulates the interaction between the ego vehicle (EV) and HVs as a zero-sum differential game \cite{wang2020infusing}. It provides a constructive method for computing the maximal control invariant set for nonlinear systems. Recent studies have successfully integrated HJ-based safety analysis into MPC frameworks as a safety filter \cite{borquez2025dualguard, leung2020infusing}. Moreover, filter-aware planning has been explored to mitigate the performance degradation caused by reactive safety interventions \cite{hu2022sharp, hu2024active}. Rather than treating safety as a post hoc correction, these methods proactively anticipate filter activations to balance task performance against potential emergency maneuvers triggered by rare but critical events. 

The strengths of CBF and HJ reachability analysis are unified within the control barrier-value function (CBVF) concept \cite{choi2021robust}. It provides a constructive and computationally tractable way to derive the maximal safe set while naturally handling bounded controls and uncertain interactions \cite{tonkens2022refining}. This perspective highlights the potential of using HJ value functions not just as reactive shielding, but also as proactive guidance for generative planning.


In this work, we present \texttt{DualShield}, a model predictive diffusion framework that realizes this potential. Unlike existing methods, \texttt{DualShield} reuses HJ value functions both to proactively guide diffusion-based trajectory generation and to construct a formal safety filter for the executed control. This dual use of the HJ value function bridges generative multimodal planning and principled safety in a unified, computationally tractable framework for interactive driving. The main contributions of this work are threefold:
\begin{enumerate}





\item We propose a unified planning and control framework, \texttt{DualShield}, that reconciles multimodal exploration with formal safety assurance. Its core mechanism repurposes HJ value functions for both proactive guidance of the denoising process and reactive shielding at execution time.

\item We introduce a tractable safety shield for multi-agent interactions. It is formulated as real-time CBVF-QP from a reusable, pre-computed pairwise HJ value functions, which apply minimal yet formal safety interventions.

\item We conduct extensive empirical validation in challenging interactive driving scenarios. The results demonstrate that \texttt{DualShield} achieves a superior balance between safety and high task efficiency under uncertainty, outperforming representative methods from different planning paradigms.
\end{enumerate}

\section{PROBLEM FORMULATION}

We consider a multi-agent setting where an EV must navigate an environment populated by $M$ HVs. Let $x \in \mathcal{X} \subset \mathbb{R}^n$ and $u \in \mathcal{U} \subset \mathbb{R}^m$ denote the state and control vectors of the EV, respectively, where $\mathcal{U}$ is a convex set. Similarly, let $x_h \in \mathcal{X}_h \subset \mathbb{R}^{n}$ and $u_h \in \mathcal{U}_h \subset \mathbb{R}^{m}$ be the state and control vectors of the $h$-th HV. The EV and HV are governed by the following  control-affine dynamics:
\begin{equation}
\label{eq:continuous_dynamics}
\dot{x} = f(x)+g(x)u, \quad \dot{x}_h = f_h(x_h)+g_h(x_h)u_h,
\end{equation} 
where $f$, $g$, $f_h$, and $g_h$ are known continuous functions characterizing the system dynamics.

Safety of the EV is defined by a requirement to avoid a failure set $\mathcal{F}$. This set is specified by a Lipschitz continuous distance function $l: \mathcal{X} \times \mathcal{X}_h \to \mathbb{R}$, such that $\mathcal{F} = \{ (x, x_h) \mid l(x, x_h) \le 0 \}$. The EV must ensure that the system state remains outside of this failure set for the entire duration of its operation.

The primary objective is to find an optimal control trajectory $u(\cdot)$ over a time horizon $[0, T]$ that minimize a performance-oriented objective functional $J_{\text{perf}}$ while ensuring safety. Crucially, this safety assurance must hold against the set of all plausible behaviors $u_h\in\mathcal{U}_h$ for all HVs, which introduces significant uncertainty. This formally defines a constrained, differential game-like optimal control problem: 
\begin{subequations}\label{eq:opt_problem}
\vspace{-4mm}
\begin{align}
\min_{u(\cdot)} \quad & J_{\text{perf}}(x,u) = \int_{0}^{T} \mathcal{L}(x, u, x_h) dt + \mathcal{L}_T(x(T)) \label{eq:to_obj_cont} \\
\text{s.t.} \quad & x(0) = x^{\text{init}}, \quad x_h(0) = x_h^{\text{init}}, \label{eq:init_cond_cont} \\
& \dot{x} = f(x)+g(x)u, \quad u \in \mathcal{U}, \label{eq:ev_dyn_cont} \\
& \dot{x}_h = f_h(x_h)+g_h(x_h)u_h, \quad u_h \in \mathcal{U}_h,  \label{eq:hv_dyn_cont} \\
& l(x, x_h) \geq 0, \quad  h\in I_0^{M-1},\label{eq:safety_cont}\\
& x\in \mathcal{X},  \label{eq:state_cont} 
\end{align}
\end{subequations}
where $\mathcal{L}$ and $\mathcal{L}_T$ are the running and terminal cost functions, respectively; and $I_0^{M-1}$ denotes $\{0,1,...,M-1\}$.

The trajectory optimization problem \eqref{eq:opt_problem}, characterized by its potentially non-convex and non-differentiable cost $\mathcal{L}$, is challenging to solve. 
The primary challenge lies in ensuring safety against the uncertain and potentially adversarial behaviors of the HVs. This transforms the problem from a standard optimal control task into a differential game, where the EV must find a strategy that is robust to the actions of all other HVs.

\section{METHODOLOGY}
\subsection{Hamilton-Jacobi Reachability Analysis}
To ensure safety, we employ HJ reachability analysis. This method computes the backward reachable set (BRS) of the failure set $\mathcal{F}$ by solving a zero-sum differential game. Compared to many forward reachability methods that assume open-loop behaviors, BRS computation in a game-theoretic setting provides tighter safety guarantees by accounting for the optimal worst-case closed-loop responses of the HVs. It is able to systematically handle nonlinear dynamics, making it particularly suitable for safe interactive driving \cite{leung2020infusing}. 

We formulate the safety problem in a relative reference frame centered at the EV and aligned with its body frame. This formulation allows the computed BRS to be reused across different dynamic scenarios. Although computationally intensive, the BRS can be pre-computed offline for efficient real-time lookup. Let $x_r$ be the relative state. It is straightforward to verify that the resulting relative dynamics also preserve a control-affine structure:
\begin{equation}
\label{eq:relative_dynamics}
\dot{x}_r = f_r(x_r, u, u_h)=f_0(x_r) + G_A(x_r)u + G_B(x_r)u_h,
\end{equation}
where $f_0(x_r)$ is the drift vector, and $G_A$ and $G_B$ are the control-input matrices for the EV and HV, respectively.

The failure set in this frame can be equivalently denoted as $\mathcal{F}_r = \{x_r \mid l_r(x_r) \le 0\}$, where $l_r$ is the distance between the EV and HV in relative coordinates. Then, the payoff function for the EV in this safety game is defined as $\mathcal{J}(x_r, u, u_h)=\min_{\tau\in [t,T_{hj}]}\ l_r(x_r(\tau))$, capturing the minimum distance from time $t$ to $T_{hj}$. We compute the optimal control of the EV to maximize this cost function with the worst-case maneuvers of the HV. This gives the value function as
\begin{equation}
V(x_r)=\sup_{u\in\mathcal{U}}\inf_{u_h\in\mathcal{U}_h}{\mathcal{J}(x_r,u,u_h)}. 
\end{equation}
This value function can be computed using dynamic programming (DP), resulting in the final value Hamilton-Jacobi-Issacs variational inequality (HJI-VI) \cite{bansal2017hamilton}: 
\begin{subequations}\label{eq:HJI-VI}
\vspace{-1mm}
    \begin{align}
\min&\{l_r(x_r)-V(x_r),\frac{\partial}{\partial t}V(x_r)+H(x_r,\nabla V(x_r))\}=0,\\
&V(x_r(T_{hj}))=l_r(x_r(T_{hj})), \ \text{for}\ t\in[0,T_{hj}],
    \end{align}
\end{subequations}
where $\nabla V(x_r)=\frac{\partial}{\partial x} V(x_r)$, and $H$ is the Hamiltonian:
\begin{equation}
\vspace{-1mm}
H(x_r,\frac{\partial}{\partial x})=\sup_{u\in\mathcal{U}}\inf_{u_h\in\mathcal{U}_h}\nabla V(x_r)f_r(x_r, u, u_h). 
\end{equation}
The BRS is given as a sublevel set of the value function:
\begin{equation}
\mathcal{B}(x_r)=\{x_r\ |\ V(x_r)\leq 0\}.
\end{equation}
It represents the set of states from which the EV cannot prevent a future collision against the worst-case maneuvers of the HV. Correspondingly, the set of provably safe states is the complement, defined as $S(x_r) = \{x_r \mid V(x_r) \geq 0\}$. The value function $V(x_r)$ thus serves as a formal safety certificate.

\subsection{Model-Based Diffusion}
We employ a model-based diffusion model to address non-differentiable objectives, leveraging its strong multimodal exploration capabilities to navigate complex interactive scenarios and avoid local optima. We formulate the problem in discrete time for clarity of exposition and ease of implementation. To ensure dynamic constraints are inherently satisfied, the diffusion process operates over the control sequence \( \mathbf{u} = \{u_{0:T-1}\} \) over a finite horizon \( N \), with a sampling period $dt=T/N$.  The corresponding state sequence \( \mathbf{x} = \{x_{1:N}\} \) is then obtained by rolling out this control sequence from an initial state $x_0$ using the discretized dynamics derived from \eqref{eq:ev_dyn_cont}.

We cast the trajectory optimization problem as sampling from a target distribution \( p^{(0)}(\mathbf{u})\) over control sequences. This distribution is designed such that the probability of any control sequence \( \mathbf{u} \) is determined by the cost of the full state-action trajectory \( \mathbf{y} = \{\mathbf{x}, \mathbf{u}\} \). 
Here, the probability is defined in a Gibbs distribution form:
\begin{equation}
\label{eq:target_dist_u}
p^{(0)}(\mathbf{u}) \propto  \exp{(-{J}(\mathbf{y})/\lambda)}, \quad \text{where} \quad \mathbf{x} = \text{Rollout}(\mathbf{u}, x_0).
\end{equation}
The total objective function $J(\mathbf{y})$ incorporates both performance objectives and soft constraints. In our formulation, this cost corresponds to the discrete-time form of the objective \eqref{eq:to_obj_cont} and state constraint penalties \eqref{eq:state_cont}.
A key advantage of this formulation is that it circumvents the need to sample from the Dirac delta distribution of dynamically feasible trajectories, since feasibility is guaranteed by the deterministic rollout adhering to dynamics constraint \eqref{eq:ev_dyn_cont}.

Our diffusion model consists of a standard \textit{forward process}, which progressively perturbs an initial control sequence \( \mathbf{u}^{(0)} \) into isotropic Gaussian noises:
\begin{equation}
\label{eq:forward}
\mathbf{u}^{(i)} \sim \mathcal{N}(\sqrt{\bar{\alpha}_i} \, \mathbf{u}^{(0)}, \sqrt{1-\bar{\alpha}_i} \, I), 
\end{equation}
where $ \bar{\alpha}_i = \prod_{k=1}^{i} \alpha_k$ with a predefined noise schedule $\alpha_k\geq0$, and $I$ is an identity matrix.

The corresponding \textit{backward process} then reverses this procedure, aiming to reconstruct \( p^{(0)}(\mathbf{u}) \) from noises. This is achieved by estimating the score function $\nabla_{\mathbf{u}^{(i)}} \log p^{(i)}(\mathbf{u}^{(i)})$, which guides the sampling towards high-likelihood regions of the target distribution over $N_d$ denoising steps. A single reverse step is then formulated as:
\begin{equation}
\mathbf{u}^{(i-1)} = \frac{1}{\sqrt{\alpha_{i}}} \left( \mathbf{u}^{(i)} + (1 - \bar{\alpha}_{i}) \, \nabla_{\mathbf{u}^{(i)}} \log p^{(i)}(\mathbf{u}^{(i)}) \right).
\label{eq:reverse}
\end{equation}

In contrast to model-free paradigms that learn a neural network to approximate the score function, we adopt a model-based approach that computes it on-the-fly \cite{pan2024model}. This allows the planner to be highly adaptive to environmental changes without requiring retraining. 
Specifically, to estimate the score via Monte Carlo approximation, we sample a set of $N_m$ candidate clean sequences \( \mathcal{U}^{(i)}=\{\mathbf{u}_{[k]}\}_{k=0}^{N_m-1} \) from the distribution:
\begin{equation}
\label{eq:sample_distribution}
\mathbf{u}_{[k]} \sim \mathcal{N}(\frac{\mathbf{u}^{(i)}}{\sqrt{\bar\alpha_{i}}},\frac{1-\bar\alpha_{i}}{\bar\alpha_{i}}I).
\end{equation}
Then, the score function \( \nabla_{\mathbf{u}^{(i)}} \log p^{(i)}(\mathbf{u}^{(i)}) \) is estimated as follows: 
\begin{equation}
\nabla_{\mathbf{u}^{(i)}} \log p^{(i)}(\mathbf{u}^{(i)}) \approx -\frac{\mathbf{u}^{(i)}}{1 - \bar{\alpha}_{i}} + \frac{\sqrt{\bar{\alpha}_{i}}}{1 - \bar{\alpha}_{i}} \, \bar{\mathbf{u}},
\label{eq:score_estimation}
\end{equation}
where the weighted mean of the optimal control sequences \( \bar{\mathbf{u}} \) is defined by:
\begin{equation}
\label{eq:average_u}
\bar{\mathbf{u}}(\mathcal{U}^{(i)}) = \frac{\sum_{\mathbf{u} \in \mathcal{U}^{(i)}} \mathbf{u} \, p^{(0)}(\mathbf{u})}{\sum_{\mathbf{u} \in \mathcal{U}^{(i)}} p^{(0)}(\mathbf{u})}.
\end{equation}
Here, the weight $p^{(0)}(\mathbf{u})$ for each control sequence \( \mathbf{u} \) is computed according to the target distribution \eqref{eq:target_dist_u}.

\subsection{The DualShield Framework}
Our proposed method, \texttt{DualShield}, unifies generative, multimodal diffusion planning with HJ safety certificate, consisting of two core components operating in synergy: a proactive, safety-guided diffusion planner that generates safe nominal trajectories, and a reactive, verifiable safety shield that certifies the executed control actions. 

\subsubsection{Proactive Safety Guidance in Diffusion Planning}
\label{sec:guided_mpd}

The uncertain and interactive nature of multi-agent driving often renders the safe portion of the state space sparse and complex. Standard sampling procedure in model-based diffusion can therefore be inefficient. To address this, we propose to embed safety awareness into the trajectory generation process directly.

The key idea is to guide the denoising process using a safety-aware objective functional $J(\mathbf{y})$. We augment the performance objective $J_{\text{perf}}$ with a safety-guided term derived from the HJ value function:
\begin{equation}
\label{eq:dualshield_cost}
J(\mathbf{y}) = J_{\text{perf}}(\mathbf{y}) + \lambda_{s} \sum_{k=0}^{N-1} \mathcal{L}(V_{\text{min},k}),
\end{equation}
where $\mathcal{L}(V_{\text{min},k}) = \gamma \max(-V_{\text{min},k}, 0),\ \gamma>0,$ is a penalty function that discourages trajectories from entering unsafe regions $\mathcal{B}(x_r)$. The term $V_{\text{min},k} = \min_{h \in I_0^{M-1}} V(x_{r,k})$ represents the minimum HJ value over all surrounding HVs, where $x_{r,k}$ is the relative state at time step $k$.

By incorporating this safety-guided objective function into the iterative denoising procedure, the diffusion planner is naturally steered towards safer regions of the trajectory space. This process yields a nominal control sequence that explores toward safe, dynamically feasible behaviors. 

\subsubsection{Receding Horizon Planning with Safety Shielding}
\label{sec:safety_shield}

We adopt a receding horizon planning scheme to deploy the model-based diffusion model in dynamic environments. To maintain computational tractability, the implementation incorporates a warm-start strategy. 

After an initial planning step with backward denoising process from a Gaussian prior, each subsequent planning cycle is initialized by adding a limited number of noise steps to the previously computed control sequence in the forward process \eqref{eq:forward}. This warm-start procedure seeds the sampler with prior solution information to accelerate convergence, reducing the required number of denoising iterations in practice, i.e., $N_{ws}< N_d$. However, even with this efficient planning scheme, the proactive safety guidance serves as a soft penalty. The planner may still generate trajectories that trade a degree of safety for higher performance.

To provide principled safety, we introduce a reactive safety shield. This shield constitutes the second, critical use of our pre-computed HJ value function. We repurpose $V(x_r)$ as a CBVF. The core principle is to enforce the forward invariance of the safe set $S = \{x_r \mid V(x_r) \ge 0\}$ by ensuring that the condition, $\dot{V}(x_r) \ge -\alpha(V(x_r))$, is satisfied at all times. Here, $\alpha(\cdot)$ is a class-$\mathcal{K}$ function \cite{ames2019control}.

Enforcing this condition in an interactive setting requires robustness against the uncertain actions of the HVs. We achieve this by considering the worst-case instantaneous effect of the HV control inputs. By expanding the time derivative of $V(x_r)$ using the control-affine relative dynamics~\eqref{eq:relative_dynamics} and minimizing over all admissible $u_h \in \mathcal{U}_h$, we obtain the following robust CBVF constraint:
$$ L_{f_0}V(x_r) + L_{G_A}V(x_r)u + \min_{u_h \in \mathcal{U}_h} [L_{G_B}V(x_r)u_h] \ge -\alpha(V(x_r)) $$
where $L_f V := \nabla V \cdot f$ is the Lie derivative. This inequality defines a half-space of the safe control $u$ for the current relative state $x_r$ with respect to the $h$-th HV.

To ensure feasibility in extreme scenarios, we relax the safety constraint using a non-negative slack variable $\epsilon$ penalized by a large weight $c_\epsilon$. This CBVF-QP problem takes the first step nominal control $u_{\text{nom}}$ from the generated sequence $\mathbf{u}_{\text{nom}}$, and computes the safe control $u_{\text{safe}}$ by minimizing the deviation from $u_{\text{nom}}$ subject to the safety constraints:
\begin{equation}
\label{eq:safety_filter_qp}
\begin{aligned}
u_{\text{safe}} = &\operatornamewithlimits{argmin}_{u \in \mathcal{U}, \epsilon \ge 0} \quad  \|u - u_{\text{nom}}\|^2 + c_\epsilon \cdot \epsilon^2 \\
\text{s.t.} \quad & L_{G_A}V(x_r) \cdot u \ge -L_{f_0}V(x_r)  - \alpha(V(x_r))  \\ 
& - \min_{u_h \in \mathcal{U}_h}[L_{G_B}V(x_r) \cdot u_h] - \epsilon, \quad \forall\ h\in I_0^{M-1}.
\end{aligned}
\end{equation}

The integration of proactive guidance and reactive certification constitutes the \texttt{DualShield} algorithm, which is detailed in Alg. \ref{alg:mpd}. This synergy ensures high-performance, multimodal exploration while providing principled safety assurance in a unified and computationally tractable manner.

\setlength{\textfloatsep}{5pt}
\begin{algorithm}[t]
    \SetKwComment{Comment}{// }{} 
    \caption{The DualShield Algorithm}
    \label{alg:mpd}
    
    \KwIn{Current state $x_k$, relative state $x_r$, HJ value function $V\;$}
    \KwParam{Sample number $N_m$, normal denoising steps $N_d$, warm-start denoising steps $N_{ws}$}
    \SetKwInOut{Persistent}{Persistent}
    \Persistent{Previous control sequence $\hat{\mathbf{u}} \gets \vect{0}\;$}
    \BlankLine 
    \If{$t = 0$}{
        Initialize noisy controls $\mathbf{u}^{(N)} \sim \mathcal{N}(\vect{0}, \mathbf{I})$\; $N_{\text{start}} \gets N_d$\;
    }
    \Else{
        Warm-start $\mathbf{u}^{(I_{ws})}$ from $\hat{\mathbf{u}}$ via forward process \eqref{eq:forward}\; $N_{\text{start}} \gets N_{ws}$\;
    }
    \BlankLine
    \For{$i = N_{\text{start}}$ \KwTo $1$}{ \label{alg:planning_start}
        Sample $N_m$ sequences $\{\mathbf{u}_{[k]}\}_{k=0}^{N_m-1}$ from \eqref{eq:sample_distribution}\;
        Compute weights $\{p^{(0)}_k\}_{k=0}^{N_m-1}$ for all candidates using safety-guided score \eqref{eq:dualshield_cost} and \eqref{eq:target_dist_u}\;
        Estimate the sequence $\bar{\mathbf{u}}$ via \eqref{eq:average_u}\;
        Perform one reverse step to get $\mathbf{u}^{(i-1)}$ using \eqref{eq:reverse}, \eqref{eq:score_estimation}, and \eqref{eq:average_u} \; \label{alg:planning_end} 
    }
    $\mathbf{u}_{\text{nom}} \gets \mathbf{u}^{(0)}$\;
    Compute control $u_{\text{safe}}$ by solving CBVF-QP \eqref{eq:safety_filter_qp}\;
    Execute control $u_{\text{safe}}$ on the EV\;
    Store $\hat{\mathbf{u}} \gets \mathbf{u}_{\text{nom}}$ for next time step\;
\end{algorithm}

\section{SIMULATION AND RESULTS}
\label{sec:simulation}

To validate the effectiveness of our proposed \texttt{DualShield} framework, we design challenging interactive driving scenarios and conduct a comprehensive comparative analysis against the baselines. 

\vspace{-1.5mm}
\subsection{Task: Unprotected Interactive U-Turn Under Uncertainty}

The EV is required to conduct a U-turn before merging into the main lane. This task is challenging as it requires navigating a non-convex space while handling significant interaction uncertainty from oncoming HVs.

\vspace{-1.5mm}
\subsection{Experimental Setup}
\subsubsection{Vehicle and Environment Modeling}
Let the state vector of the EV be $x = [p_x, p_y, \theta, v]^T$ and the control vector be $u=[w, a]^T$, where $w$ is the yaw rate and $a$ is the longitudinal acceleration. Similarly, the state vector of the HV is $x_h = [p_{x,h}, p_{y,h}, \theta_h, v_h]^T$ with control vector $u_h = [w_h, a_h]^T$ defined analogously. Their world-frame dynamics are as follows:
\begin{align*}
    \dot{p}_x &= v \cos(\theta), \quad \dot{p}_y = v \sin(\theta), \quad \dot{\theta} = w, \quad \dot{v} = a, \\
    \dot{p}_{x,h} &= v_h \cos(\theta_h), \quad \dot{p}_{y,h} = v_h \sin(\theta_h), \quad \dot{\theta}_h = w_h, \quad \dot{v}_h = a_h.
\end{align*}

We denote the relative state as $x_r = [p_{x,r}, p_{y,r}, \phi_r, v, v_h]^T$ in the body frame of EV, where $p_{x,r},\ p_{y,r}$ denote the relative position of the HV, $\phi_r = \theta_h - \theta$ denotes relative orientation, and $v, v_h$ are absolute speeds of the EV and HV, respectively.

The transformation from world to body frame is given by:
$$
\begin{bmatrix} p_{x,r} \\ p_{y,r} \end{bmatrix} = 
\begin{bmatrix} \cos(\theta) & \sin(\theta) \\ -\sin(\theta) & \cos(\theta) \end{bmatrix}
\begin{bmatrix} p_{x,h} - p_x \\ p_{y,h} - p_y \end{bmatrix}
$$
By taking the time derivative of each component of $x_r$, we obtain the relative dynamics: 
$$
\dot{x}_r = 
\underbrace{
\begin{bmatrix}
-v + v_h \cos(\phi_r) \\
v_h \sin(\phi_r) \\
0 \\
0 \\
0
\end{bmatrix}
}_{f_0(x_r)}
+
\underbrace{
\begin{bmatrix}
p_{y,r} & 0 \\
-p_{x,r} & 0 \\
-1 & 0 \\
0 & 1 \\
0 & 0
\end{bmatrix}
}_{G_A(x_r)}
\underbrace{
\begin{bmatrix}
w \\ a
\end{bmatrix}
}_{u}
+
\underbrace{
\begin{bmatrix}
0 & 0 \\
0 & 0 \\
1 & 0 \\
0 & 0 \\
0 & 1
\end{bmatrix}
}_{G_B(x_r)}
\underbrace{
\begin{bmatrix}
w_h \\ a_h
\end{bmatrix}
}_{u_h}.
$$

Simulations are conducted in a $1:4$ scaled environment mirroring a real-world intersection. There are two HVs and 20 static obstacles in the setting. The lane width is set to $1.5\, \text{m}$. Both the EV and HVs are modeled with a rectangle footprint of $1.0\, \text{m}$ in length and $0.4\, \text{m}$ in width. For safety analysis, this footprint is approximated by a dual-circle model, which consists of two circles, each with a radius of $0.3\, \text{m}$. The centers of these circles are located $0.25\, \text{m}$ ahead of and behind the geometric center along its longitudinal axis. Static obstacles, representing lane dividers, are modeled as circles with a $0.1\, \text{m}$ radius. To enhance computational efficiency, we treat the EV as a point and inflate the radii of all other objects accordingly. Additionally, only the three nearest static obstacles are considered in the safety shield. For the EV, the angular velocity $\omega \in [-\pi/3, \pi/3]\, \text{rad/s}$ and acceleration $a \in [-1, 1]\, \text{m/s}^2$; whereas for all the HVs, $\omega_h \in [-\pi/18, \pi/18]\, \text{rad/s}$ and $a_h \in [-1, 1]\, \text{m/s}^2$.

We use constant velocity prediction for the planning methods. However, to model the uncertainty, the HVs are actually programmed with three distinct interactive patterns, randomly selected at the start of each trial:
\begin{itemize}
\item \textbf{Cooperative:} The HV yields to the preceding EV, maintaining a safe following distance using the intelligent driver model (IDM). 
\item \textbf{Oblivious:} The HV ignores the EV and maintains its pre-defined lane-following velocity.
\item \textbf{Adversarial:} The HV accelerates to contest the right-of-way at its max acceleration, creating a worst-case threat for the EV.
\end{itemize}

\subsubsection{Objective Function and Parameter Setting}
The performance objective, $J_{\text{perf}}$, encodes desirable driving behaviors, requiring the EV to execute a U-turn from an initial state $x_0 = [2, 0.7, \pi, 0.5]^T$ to a target driving state $x_g = [p_{x,g},p_{y,g},\theta_{g},v_{g}]^T=[2, -0.7, 0, 0.5]^T$. It is a weighted sum over a planning horizon of $N$ steps:
\begin{equation}
\vspace{-2mm}
J_{\text{perf}} = \sum_{k=0}^{N-1} \left( w_{\text{goal}}J_{\text{goal}}(x_k) + w_{\text{reg}}J_{\text{reg}}(x_k, u_k) \right),
\label{eq:cost_full}
\end{equation}
where the goal-tracking cost $J_{\text{goal}}(x_k) = \|x_k - x_g\|^2_Q$ penalizes the weighted squared error between the state $x_k$ and $x_g$, and weighting matrix is $Q = \text{diag}(0, 20, 5, 1)$. Notably, there is no penalty on the longitudinal position $p_x$, as the objective is to merge into the target lane and subsequently engage in lane-keeping, rather than reach a specific waypoint. The non-differentiable regularization cost $J_{\text{reg}}=J_{\text{rule}}+J_{\text{boundary}}+J_{\text{spin}}$ encourages smooth and rule-compliant driving. It consists of three components: $J_{\text{rule}}=\gamma_{\text{turn}} \max(0, p_y) \cdot \max(0, \cos(\theta))$ penalizes driving in the wrong direction; $J_{\text{boundary}}=\gamma_{b} (\max(0, p_y - y_{\max})^2 + \max(0, y_{\min} - p_y)^2)$ is a penalty for violating the lateral boundaries at $[y_{\min}, y_{\max}]=[-1.5,1.5]$; and $J_{\text{spin}}=\gamma_{\text{spin}}\omega^2 \exp(-c_v v^2)$ penalizes dry steering behaviors.
All weights are empirically tuned for desired behaviors: $\gamma_{\text{spin}}=1$, $c_v=5$, $\gamma_{\text{turn}}=50$, $\gamma_{b}=20$, and $c_v=5$. The coefficients for the HJ safety term~\eqref{eq:dualshield_cost} is set to $\gamma=10$ and $\lambda_s=1$.

\vspace{-2mm}
\subsection{Baselines for Comparison}
\vspace{-1.5mm}
\label{ssec:baselines}
To comprehensively validate the superiority of our framework, we benchmark its performance against three baselines, each meticulously selected to represent a dominant and distinct planning paradigm:
\begin{itemize}
\item \textbf{Model-Based Diffusion (MBD) \cite{pan2024model}:} As a state-of-the-art representative of generative planners, MBD utilizes a diffusion model for multimodal  trajectory generation, enforcing safety through a distance-based penalty score function. For a fair comparison, it is configured with the same receding horizon and warm-start strategy as our method.
\item \textbf{Nonlinear MPC (NMPC):} A standard NMPC handles obstacle avoidance with nonlinear constraints using CasADi and IPOPT solver. Its cost function is configured to be analogous to ours for a fair comparison.
\item \textbf{DualGuard-MPPI \cite{borquez2025dualguard}:} This baseline represents the paradigm of sampling-based planners with strong safety considerations. It employs the same HJ value function as a rejection sampler for safe rollouts and as a trigger for a switching-based safety filter.
\end{itemize}

\subsection{Implementation and Evaluation}
\label{ssec:implementation_metrics}

\subsubsection{Implementation Details}

The value functions for relative dynamics are pre-computed using the \texttt{hj\_reachability} toolbox\footnote{\url{https://github.com/StanfordASL/hj_reachability}}, which implements the methods described in \cite{bansal2017hamilton}. The distance function $l(x_r)$ for failure set $\mathcal{F}_r$ is designed as $l_r(x_r)=x_r^2-r_s^2$, where $r_s$ is set to $0.6$ and $0.4$ for HVs and static obstacles, respectively. We use a $100\times100\times64\times8\times8$ grid for the 5D relative state $(p_{x,r}, p_{y,r}, \psi_r, v, v_h)$ over the domain $[-8,8]\times[-8,8]\times[0,2\pi]\times[0,4]\times[0,4]$ (m, m, rad, m/s, m/s). $T_{hj}$ is set to $1\,\text{s}$. The pre-computation takes approximately $3\, \text{h}$ on a workstation equipped with 2.60 \,\text{GHz} Intel Xeon Platinum 8358P CPU. 

Our simulation environment is developed in Python, leveraging the JAX library for high-performance numerical computations and automatic differentiation. Key hyperparameters include $N_d=100$ normal denoising steps, $N_{ws}=5$ warm-start denoising steps, and a sample size of $N_m=2000$. Our planning horizon is set to $N=50$ with a sampling period of $dt=0.1\,\text{s}$. The total simulation duration is $10\, \text{s}$. The penalty weight in \eqref{eq:safety_filter_qp} for the slack variable is set to $c_\epsilon=10^8$.

To ensure statistical significance, we perform 100 simulation trials. These are structured as 10 trials for each of the 10 randomly generated scenario configurations. For each configuration, the initial speeds of HVs are sampled from $[0.5, 2.0]\, \text{m/s}$, and its interactive behavior is chosen randomly.

\begin{figure}[!t]
    \centering \hspace{-4mm}
    \includegraphics[scale=0.325]{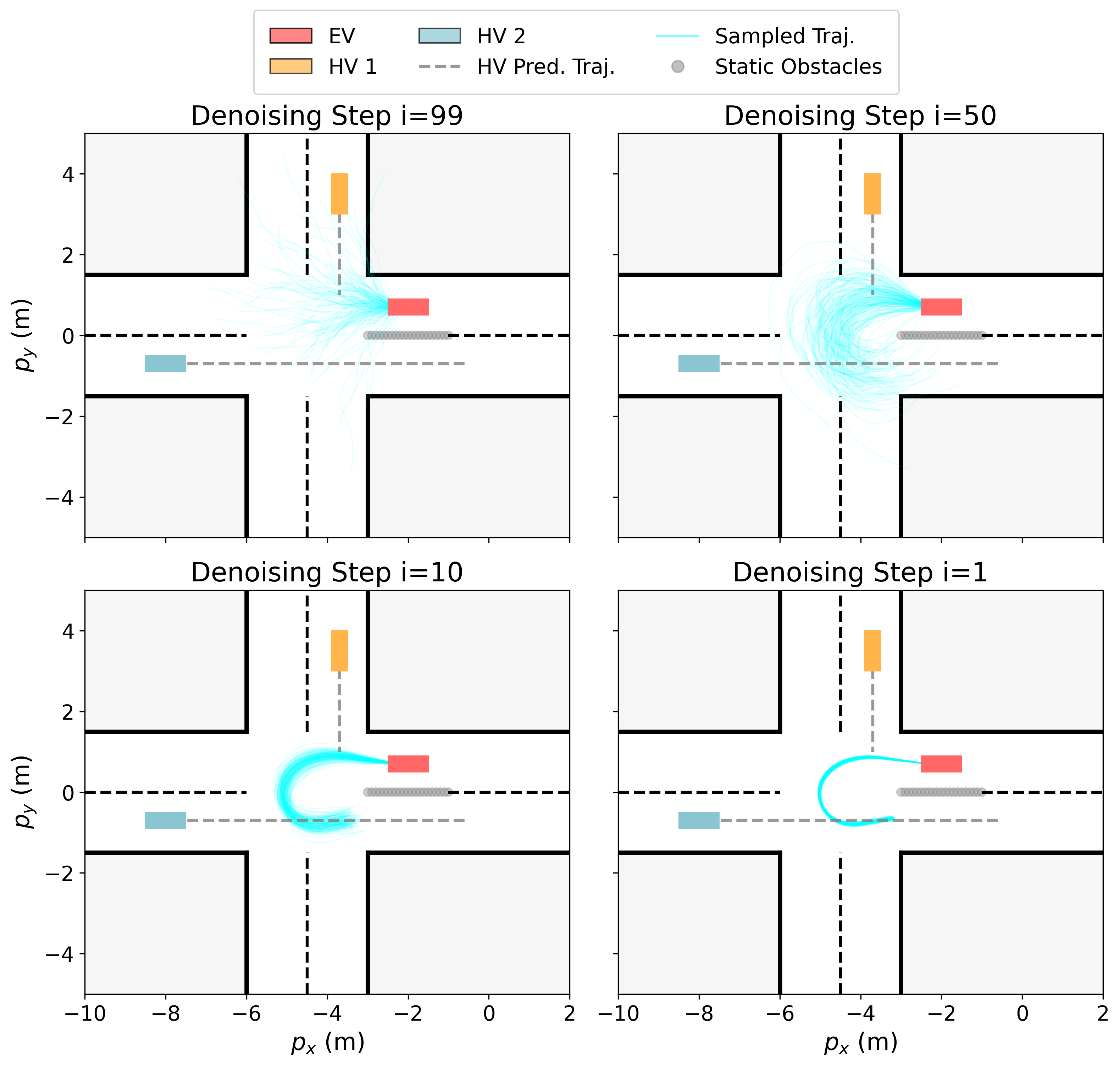}
    \vspace{-1mm}
    \caption{Visualization of the safety-guided denoising process at different steps. The planner begins with a cloud of noisy trajectories ($i=99$). As the process continues, the safety-guided score steers the distribution until it converges to a high-quality, safe trajectory distribution ($i=1$). }
    \label{fig:denoising_process}
    \vspace{-1mm}
\end{figure}

\subsubsection{Evaluation Metrics}
We evaluate the performance based on safety, mission success, and comfort:
\begin{itemize}
    \item \textbf{Safety:} Collision rate $P_c$ (\%) and the minimum distance $l_{r,min}\ (\text{m})$ between EV and HVs/static obstacles. A value of $l_{r,min} \leq 0$ indicates a collision.
    \item \textbf{Mission Success:} Success rate $P_s$ (\%) and mission completion time $T_m$ (s). A trial succeeds if the EV enters and remains in the target lane-keeping state for 5 consecutive steps (defined by lateral error $|p_{y}-p_{g,y}|\leq 0.2~\text{m}$, heading error $|\theta-\theta_{g}|\leq \pi/3$, and speed $|v|\ge 0.2~\text{m/s}$). The completion time is recorded at the onset of this stable period.
    \item \textbf{Comfort:} Average control jerk $j$ (m/s³).
    \item \textbf{Computation Efficiency:} Average computation time $T_c$ (s) per planning step.
\end{itemize}

\begin{table}[t]
\caption{Performance Comparison Among Different Methods.}
\label{tab:single_trial_results}
\centering
\scriptsize
\setlength{\tabcolsep}{4pt}
\begin{tabular}{@{}l l c c c c c@{}}
\toprule
\textbf{Method} & \textbf{$P_s$}\,$\uparrow$ & \textbf{$P_c$}\,$\downarrow$ & \textbf{$l_{r,min}$} (m)\,$\uparrow$ & \textbf{$T_m$} (s)\,$\downarrow$ & \textbf{$j$} (m/s$^{3}$)\,$\downarrow$ & \textbf{$T_c$} (s)\,$\downarrow$ \\
\midrule
MBD~\cite{pan2024model}              & {90\%} & {10\%} & {0.16}  & 4.3   & 3.10 & 0.41 \\
NMPC                          & {0\%}   & {0\%} & 0.28                 & –     & 0.10 & 0.38 \\
DualGuard‐MPPI~\cite{borquez2025dualguard} & {80\%}   & {0\%} & 0.24                & 7.39   & 6.40 & 0.24 \\
\textbf{DualShield (Ours)}    & {100\%} & {0\%} & 0.26       & 4.7 & 2.78 & 0.78 \\
\bottomrule
\end{tabular}
\end{table}

\vspace{-2.5mm}
\subsection{Results and Analysis}

\subsubsection{Comparative Performance Analysis}
\begin{figure}[thpb]
\centering
\includegraphics[scale=0.35]{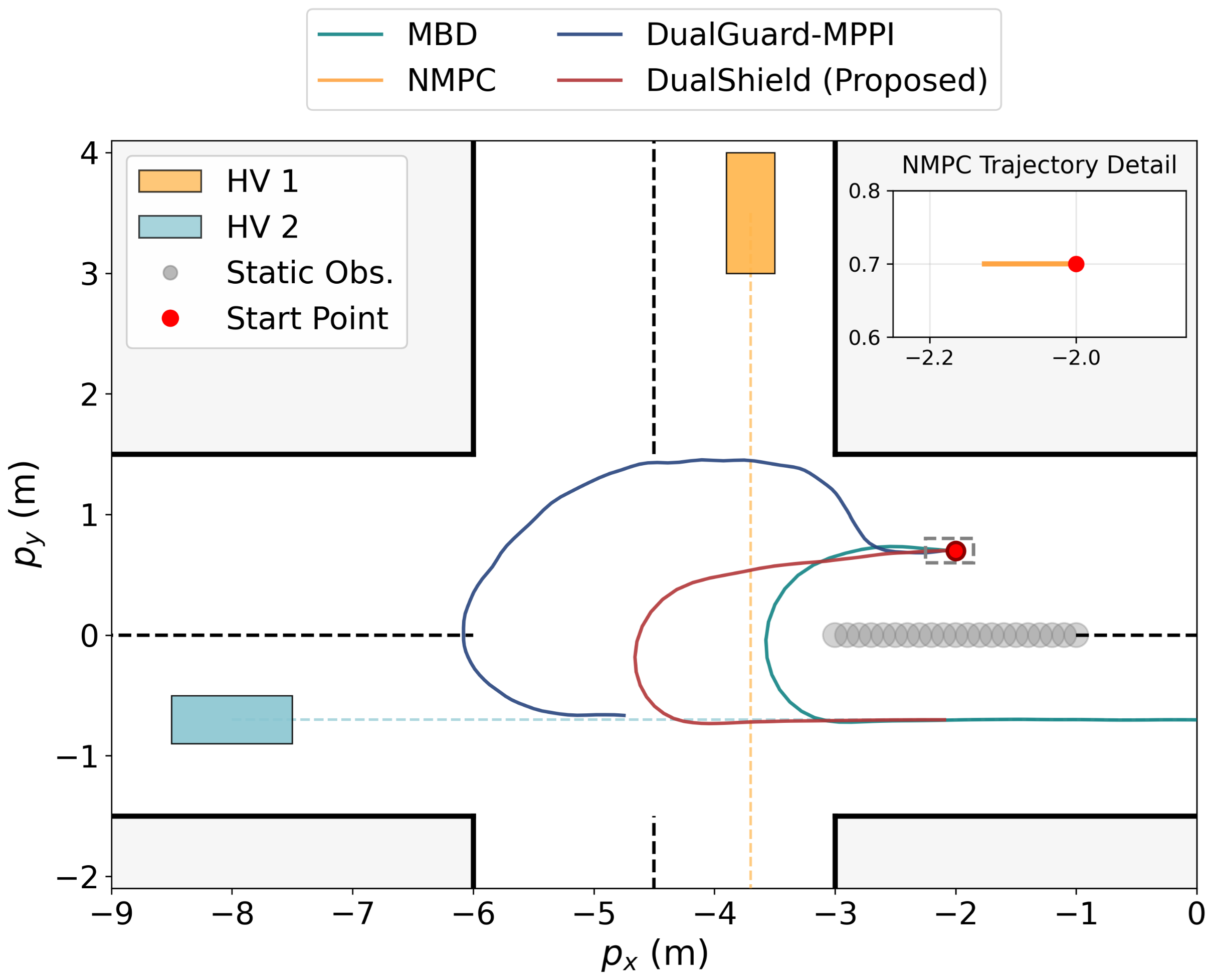}
\vspace{-3mm}
\caption{Comparison of closed-loop trajectories in a representative scenario with adversarial HV1 and HV2.
(a) MBD: Collides with the HV1 due to reliance on soft penalties.
(b) NMPC: Fails by getting trapped in a local minimum.
(c) DualGuard-MPPI: Ensures safety but results in an inefficient, hesitant trajectory.
(d) \texttt{DualShield}: Achieves a safe and efficient trajectory by proactively navigating the interaction (see Fig. \ref{fig:six_frame_subplot_with_brt} for the snapshots of dynamic interaction).}
\label{fig:closed_loop_comparison}
\vspace{-1mm}
\end{figure}

We now present a quantitative comparison of \texttt{DualShield} against the baseline methods, with results summarized in Table \ref{tab:single_trial_results}. Representative closed-loop trajectory comparisons are shown in Fig. \ref{fig:closed_loop_comparison}. While MBD achieves a high success rate (90\%), its 10\% collision rate underscores the fundamental risk of relying solely on distance-based soft collision penalties with inaccurate predictions. When the adversarial HV1 accelerates and deviates from the simple constant-velocity prediction, the trajectory of MBD becomes unsafe, leading to collisions. NMPC fails to complete the task (0\% success), highlighting its vulnerability to local minima in this highly non-convex planning landscape. DualGuard-MPPI ensures safety (0\% collision) but does so at a significant cost to performance, exhibiting the longer average completion time and higher control jerk than our method. 
This underscores the performance limitations of relying on local perturbation-based sampling. 

\begin{figure}[thpb]
\centering
\vspace{-2mm}
\includegraphics[scale=0.18]{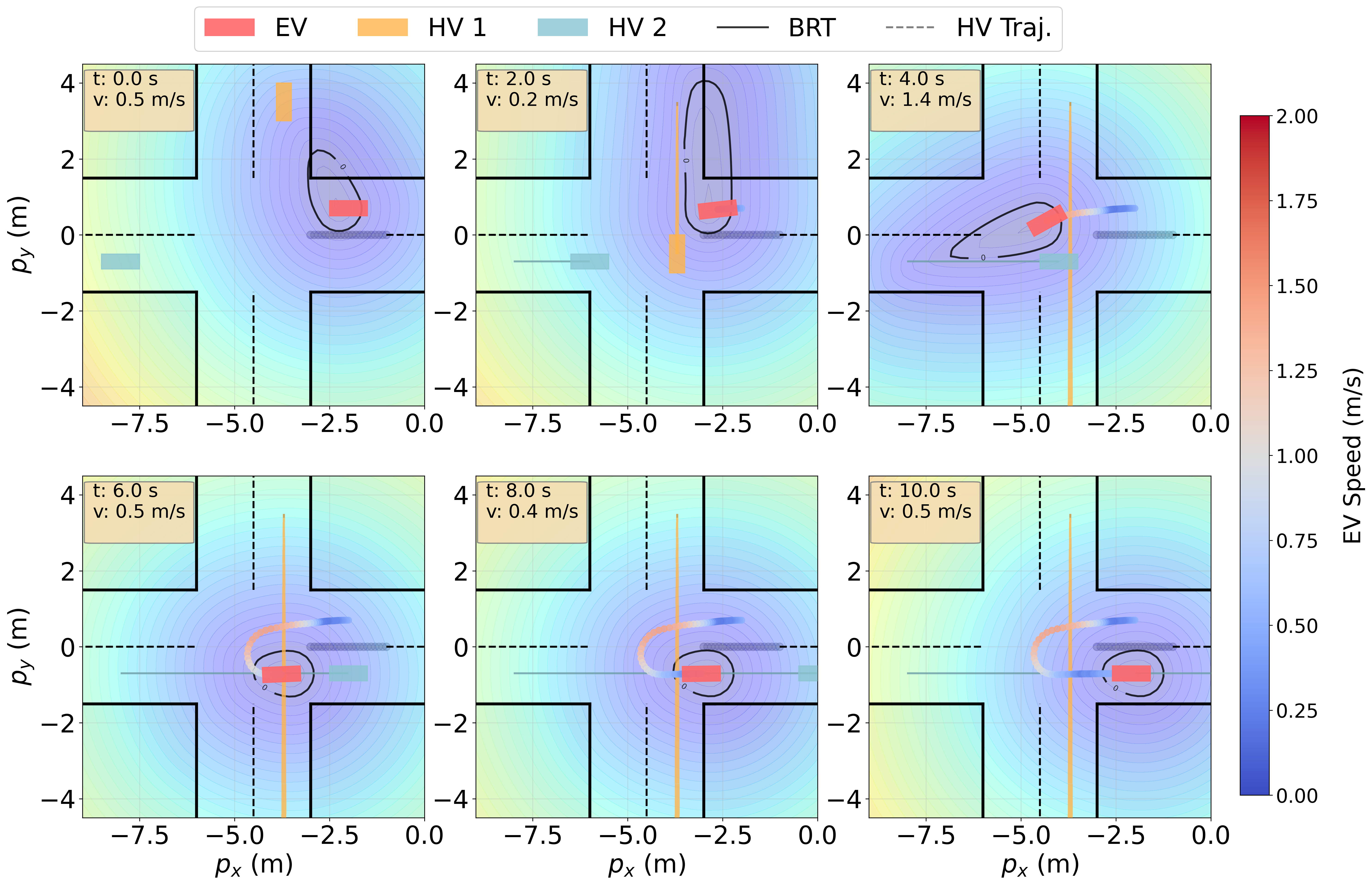}
\caption{Snapshots of \texttt{DualShield} executing a tactical U-turn. It first yields to the adversarial HV1, then accelerates to merge safely behind the adversarial HV2, ultimately achieving a stable lane-keeping state in the target lane. This safe maneuver is achieved despite relying on a simple constant velocity prediction for both HVs, as the underlying reachability analysis accounts for the full range of their potential actions. The contours depict the HJ value function relative to the most threatening HV, with the black line marking the BRT boundary.}
\label{fig:six_frame_subplot_with_brt}
\vspace{-1mm}
\end{figure}

In contrast, \texttt{DualShield} is the only method that achieves a perfect success rate (100\%) while avoiding collisions. It simultaneously attains the shortest completion time among all safe planners, which indicates the effectiveness of the proactive guidance of HJ value functions toward safe and high-performance regions. It accounts for the entire set of possible future maneuvers from the HVs, including worst-case accelerations. The result demonstrates the robustness of the DualShield in ensuring safety, even with a simple constant-velocity model for planning. We acknowledge that the current implementation is slower than the baselines, mainly due to the large number of value function queries performed during sampling and safety shielding. This computational cost can be mitigated through GPU-parallel sampling and a more efficient, batched and just-in-time (JIT) compiled value-query pipeline.
A sequence of snapshots from a representative trial, presented in Fig. \ref{fig:six_frame_subplot_with_brt}, visualizes this effective interplay between planning and safety.

\subsubsection{Analysis of the Safety-Guided Denoising Process}
To understand how \texttt{DualShield} generates high-quality trajectories, we visualize the trajectory distribution at different stages of the iterative denoising process within a single planning cycle, shown in Fig. \ref{fig:denoising_process}. The planner starts with a diverse set of noisy candidate sequences sampled from a Gaussian prior. In the early stages of denoising, the trajectories are chaotic and explore a wide area of the state space.
As the reverse diffusion process progresses, the effect of our HJ value guidance becomes apparent. It effectively penalizes candidate trajectories that would enter unsafe regions, steering the distribution away from them. Simultaneously, the goal and regularization terms guide the candidates toward the desired task goal. This guided evolution rapidly collapses the distribution from a noisy, high-entropy state to a low-entropy distribution concentrated around a safe and dynamically feasible optimal trajectory. This demonstrates the efficiency of our approach in navigating non-convex optimization landscapes with complex non-differentiable objective function.

\subsubsection{Flexible Multimodal Behaviors under Uncertainty}

\begin{figure*}[tp]
\centering
\includegraphics[scale=0.29]{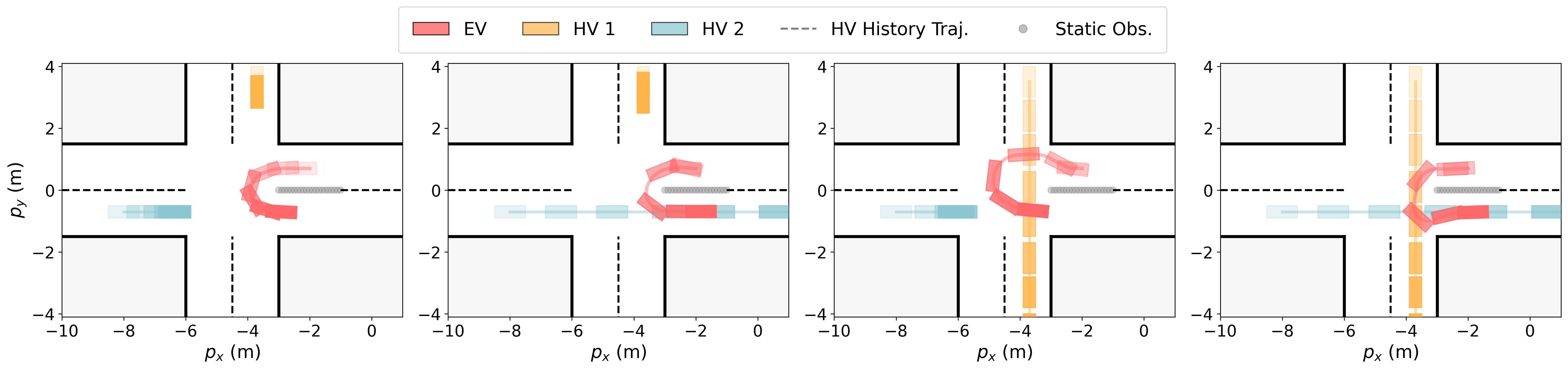}
\vspace{-2.5mm}
\caption{Flexible multimodal planning of \texttt{DualShield} in varying multi-agent interactions. 
    (a) When both HVs are cooperative, \texttt{DualShield} plans an assertive merge. 
    The planner then demonstrates robust adaptation to mixed-intent scenarios: (b) It safely navigates a contested space created by a yielding HV1 and an adversarial HV2. (c) it adeptly handles the mirrored case  with an adversarial HV1 and a yielding HV2. 
    (d) Finally, when both HVs act adversarially, the planner correctly identifies the high risk and switches to a safe, defensive yielding maneuver, robustly selecting the appropriate behavioral mode. (The opacity indicates temporal progression, with current states at full opacity and past positions fading progressively.)}
\label{fig:multimodal_response}
\vspace{-6mm}
\end{figure*}

Fig. \ref{fig:multimodal_response} showcases the multimodal planning capability of \texttt{DualShield}.
This ability to select the optimal strategy from a rich, safety-certificated distribution of behaviors highlights how \texttt{DualShield} unifies discrete strategic reasoning with continuous trajectory optimization. It leverages the exploratory power of generative models while anchoring every decision in principled safety assurance.


\vspace{-1mm}
\section{CONCLUSION}
\vspace{-1.5mm}
In this paper, we presented \texttt{DualShield}, a unified planning and control framework that resolves the core tension between multimodal exploration and formal safety in dynamic interactive driving. 
By repurposing HJ value functions for both proactive guidance and reactive shielding, \texttt{DualShield} robustly handles uncertain interactions with principled safety. Our simulation results demonstrate that this dual-use architecture enables high mission success rates and efficient trajectories while maintaining safety, a balance that proved challenging for baseline methods. 
Looking ahead, a key research avenue is the online approximation of HJ values via reinforcement learning, removing the dependency on pre-computed values and allowing for greater generalization.
\vspace{-2mm}


\bibliographystyle{IEEEtran}
\bibliography{ref}

\end{document}